\documentclass{article}

\usepackage[nonatbib,preprint]{neurips_2020}

\usepackage[utf8]{inputenc} 
\usepackage[T1]{fontenc}    
\usepackage{hyperref}       
\usepackage{url}      
\usepackage{amsmath}
\usepackage{wrapfig}
\usepackage{booktabs}       
\usepackage{amsfonts}       
\usepackage{nicefrac}       
\usepackage{microtype}      
\usepackage{multicol}
\usepackage{array}
\usepackage{graphicx}
\usepackage{multirow}
\usepackage{color}

\title{A Crystal-Specific Pre-Training Framework for Crystal Material Property Prediction}

 \author{%
   Haomin Yu$^1$, Yanru Song$^2$, Jilin Hu$^{2}$, Chenjuan Guo$^{2}$, Bin Yang$^{2}$ \\
   $^1$Department of Computer Science, Aalborg University\\
    $^2$School of Data Science 
\& Engineering, East China Normal University\\
haominyu@cs.aau.dk, 51255903111@stu.ecnu.edu.cn, \{jlhu, cjguo, byang\}@dase.ecnu.edu.cn
}

\begin{document}

\maketitle

\begin{abstract}

Crystal property prediction is a crucial aspect of developing novel materials.  However, there are two technical challenges to be addressed for speeding up the investigation of crystals. First, labeling crystal properties is intrinsically difficult due to the high cost and time involved in physical simulations or lab experiments. Second, crystals adhere to a specific quantum chemical principle known as periodic invariance,  which is often not captured by existing machine learning methods. To overcome these challenges, we propose the crystal-specific pre-training framework  for learning crystal representations with self-supervision. The framework designs a mutex mask strategy for enhancing representation learning so as to alleviate the limited labels available for crystal property prediction. Moreover, we take into account the specific periodic invariance in crystal structures by developing a periodic invariance multi-graph module and periodic attribute learning within our framework.  This framework has been tested on eight different tasks. The experimental results on these tasks show that the framework achieves promising  prediction performance  and is able to outperform recent strong baselines. 

\end{abstract}
\vspace{-10pt} 
\section{Introduction}
 \vspace{-5pt}
A crystal is a solid material with a specific and repeating arrangement of atoms, molecules, or ions that form a highly ordered internal structure \cite{monaco2002fundamentals}. 
By virtue of their internal structure, crystals exhibit unique physical and chemical properties that make them valuable in numerous industrial applications. For example, semiconductor crystals contribute to industrial electronic devices.  Beyond their technological relevance, crystals also hold biological significance and have been utilized in the development of medicines and medical treatments \cite{strydom2017hemilabile}. In light of the significance of crystals, this work focuses on mining crystal structures to achieve accurate crystal property prediction.

In recent years, the fast development of machine learning methods has attracted tremendous interest in accelerating materials design. 
Traditional approaches to researching and developing crystalline materials involve conducting numerous experiments, which can be time-consuming and resource-intensive\cite{abola2000automation,wei2019machine}. Moreover, such approaches are often limited by experimental conditions and theoretical constraints. Then, density functional theory~(DFT) calculations have emerged as a valuable tool for characterizing material properties and discovering new materials \cite{curtarolo2013high}. However, despite the success of high-throughput DFT in materials discovery, it is still constrained by its relatively high computational cost \cite{ward2017including}, which limits its utility in predicting properties of novel materials~\cite{park2020developing}. Most recently, graph neural networks (GNNs)~\cite{scarselli2008graph} have emerged as promising methods to model graph-structured crystal data and achieved competitive crystal property prediction accuracy in mining graph-structured dependencies.

However, predicting crystal properties remains difficult due to the following challenges. (1) \textbf{Limited number of labeled crystal data}. 
Although machine learning models are promising in predicting crystal properties, it requires a tremendous amount of labeled crystal data to train the models. Yet, labeling crystal properties is always challenging since it is labor-intensive and calls for a large amount of time and effort in physical simulations or laboratory experiments. 
(2) \textbf{Quantum chemical constraints}. 
The crystal structure satisfies several quantum chemical constraints by genuine, e.g., including E(3) invariance (i.e., translation, rotation and reflection invariances) and periodic invariance, so it is essential to consider these constraints when designing crystal property prediction machine learning models. However, most existing learning methods mainly focus on studying E(3) invariance while ignoring the periodic invariance, which limits their ability in  predicting crystal properties.

To contend with the above challenges, we propose a mutex masked pre-training (MMPT) framework  while ensuring periodic invariance. \textit{To reduce the amount of labeled crystal data}, we try to leverage a large amount of unlabeled crystal data and propose a self-supervised framework with the help of mutex masking (cf. Section \ref{mutex_mask}), which provides two mutually exclusive views to pre-train a property prediction network, thus reducing the amounts of labeled data.  
 \textit{To meet the quantum chemical constraints of the crystal}, in particular the periodic invariance, we design two periodic-specific modules, which include a periodic invariance multi-graph (PIMG) module  and a periodic attribute learning (PAL) module. The PIMG module employs a novel multi-graph attention network to capture complex and subtle structural patterns by leveraging the structural information of crystals. The PAL module incorporates a crystal periodic structure, which can lead the model to capture the underlying physical properties and chemical behaviors of crystals. To the best of our knowledge, it is the first work that utilizes explicit periodic-specific features in a self-supervised pre-training framework. The contributions of this work can be summarized as follows:
(1) We propose a novel mutex masked strategy to obtain two mutually exclusive views to enable effective self-supervised pretraining. This approach is able to reduce the amount of labeled data for property prediction.  (2) We design a periodic invariance multi-graph (PIMG) module that leverages the structural patterns of crystals to enhance the expressiveness of the learned representations while following the periodic invariance. It allows us to capture complex and subtle structural patterns. (3) We customize a periodic attribute learning module for our self-supervised framework, which explicitly designs periodic attributes while adhering to crystal quantum chemistry principles. (4) We conduct experiments on eight crystal datasets, and the experimental results demonstrate the effectiveness of our proposed MMPT.

 \vspace{-5pt}
\section{Related Work}
  \vspace{-5pt}
    \subsection{ Material Property Prediction} 

    In recent years, there has been a significant increase in the application of machine learning techniques, specifically graph neural networks (GNN), for predicting material properties. GNN-based methods for predicting material properties can be divided into two categories: 2D molecular graphs \cite{yang2019analyzing,coley2019graph,unke2019physnet} and 3D molecular graphs \cite{schutt2017schnet,xie2018crystal,chen2019graph,gasteigerdirectional}. 
    Since the property of crystal materials strongly relies on their 3D information, we focus on methods that can utilize 3D molecular graphs. 
    Sch\"utt et al. \cite{schutt2017schnet} proposed a neural network specifically designed to respect essential quantum chemical constraints by designing a continuous-filter convolutional layer. Xie et al. \cite{xie2018crystal} developed a crystal graph convolutional neural networks framework to directly learn the material properties. Chen et al. \cite{chen2019graph} proposed graph networks with global state attributes (MegNet) to achieve property prediction.  DimeNet \cite{gasteigerdirectional} and DimeNet++ \cite{gasteiger2020fast} allow GNNs to incorporate directional message passing via bond angles, which further improve property prediction. ALIGNN \cite{choudhary2021atomistic} performs message passing on both the interatomic bond graph and its line graph. 
    However, all these methods do not explicitly consider the dynamic periodic pattern of crystals. To this end, Matformer~\cite{yanperiodic} introduces a graph construction method, which encodes lattice information into crystal graphs by adding self-connecting edges. 
    However, this method cannot capture periodic patterns adaptively and is trained in an end-to-end supervised learning fashion. In our work, we introduce the periodic invariance multi-graph module and periodic attribute learning module to capture periodic patterns adaptively in a self-supervised learning framework.
    %
    




    \subsection{Self-Supervised Learning for Molecular Representation Learning} 

In recent years, self-supervised learning (SSL) techniques have garnered increasing interest for their ability to learn meaningful representations from unlabeled data \cite{liupre, hamilton2017inductive}. SSL can be mainly divided into two categories \cite{liu2022molecular}: contrastive SSL and generative SSL. \textbf{Contrastive SSL} aims to learn meaningful representations by maximizing the similarity between pairs of augmented samples. Several studies~\cite{you2020graph,you2021graph, xu2021self,liupre,stark20223d} have proposed variants of contrastive learning-based methods to improve the performance of molecular properties prediction. 
However, all those contrastive learning methods require a bunch of negative samples for each molecule, which requires additional computation and memory resources and is a bottleneck in some applications. 
As a specific type of contrastive SSL, the Barlow Twins~\cite{zbontar2021barlow} approach has been proposed to learn useful data representations without requiring negative samples, and Magar et al.~\cite{magar2022crystal} proposed a Crystal Twins that learns representations by utilizing Barlow Twins, forcing graph latent embeddings of augmented instances obtained from the same crystalline system to be similar. However, although this approach can encourage the model to learn discriminative features, it may inadvertently neglect to introduce some periodic patterns and certain structural information.
\textbf{Generative SSL} \cite{liu2022molecular} focuses on reconstructing the original input information \cite{liupre, hamilton2017inductive,hustrategies,hu2020gpt,liu2020structured}. Generative SSL can be categorized into various types, including variational auto-encoder (VAE) \cite{liupre}, auto-encoder \cite{hou2022graphmae}, generative adversarial networks (GAN) \cite{creswell2018generative}, flow-based models \cite{satorrasn}, and diffusion-based models \cite{hoogeboom2022equivariant}. Recently, based on generative SSL, some methods have considered the success of the masking technique and integrated it into the SSL framework. Hou et al. \cite{hou2022graphmae} proposed focusing on feature reconstruction with both a masking strategy and scaled cosine error by utilizing an autoencoder method. Different from previous works,  our proposal considers both contrastive (i.e., Barlow Twins) and generative SSL (i.e., autoencoder) with the help of a novel mutex masking strategy.
The autoencoder framework emphasizes the reconstruction and feature preservation, while the Barlow Twins approach encourages the learning of discriminative features. This combination can provide a comprehensive understanding of the underlying structure.   Furthermore, through reconstructing masked  data, the model can learn the representations that take into account surrounding atoms and their relationships, rather than learning individual atom representations.

 \vspace{-5pt}
\section{Preliminaries}
 \vspace{-5pt}
\subsection{Crystal Representation}

\label{cryrep}
 A crystal can be completely specified by three vectors $\mathbf{C}=(\mathbf{A}, \mathbf{X}, \mathbf{L})$,  which is built up by repetitive \begin{wrapfigure}{r}{0.4\columnwidth}
 \vspace{-10pt}
     \includegraphics[width=0.4\textwidth]{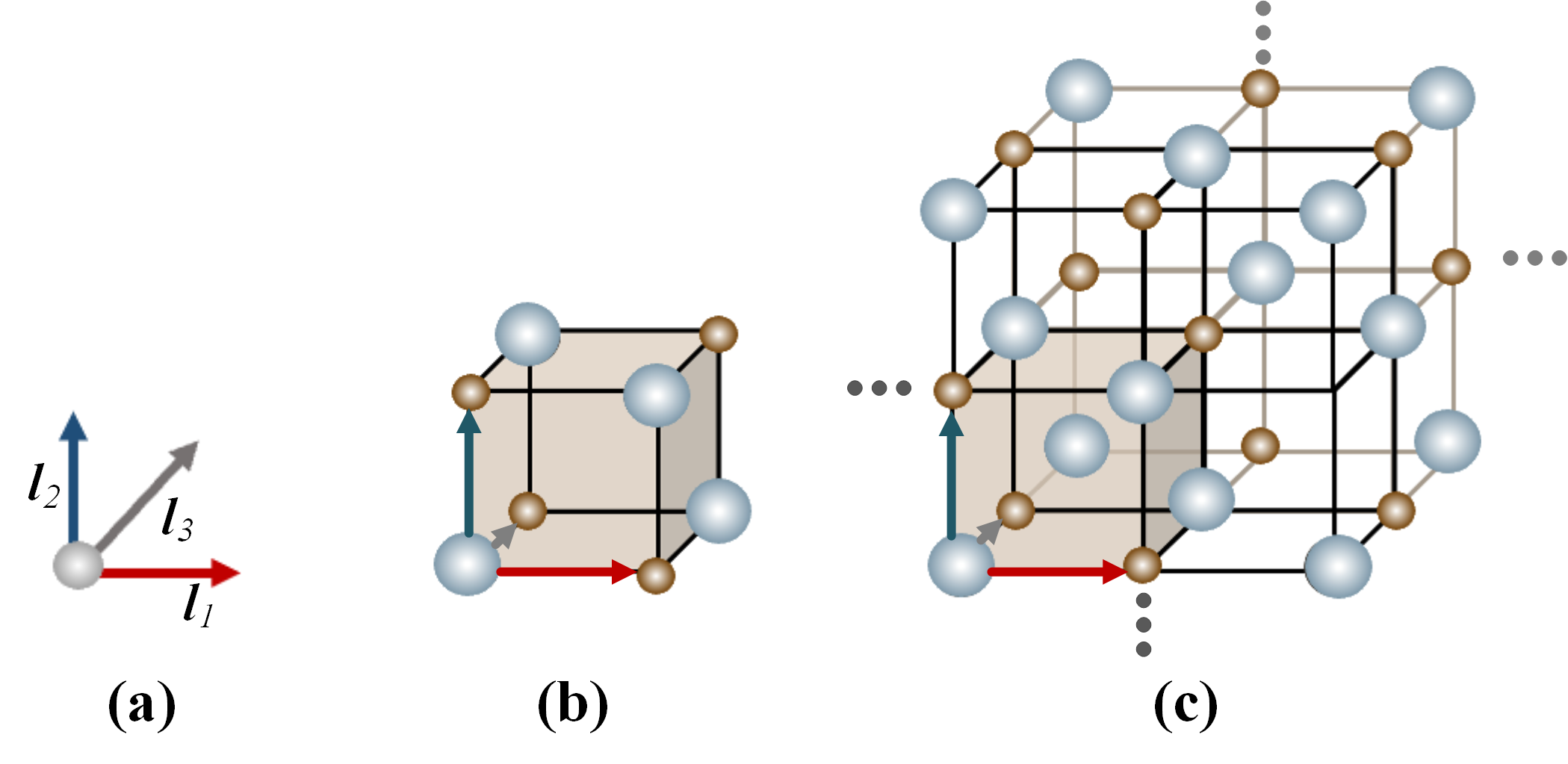}
    \caption{The illustration of crystal structure. (a) shows the periodic lattice. (b) shows a unit cell. (c) represents an infinite periodic crystal structure.}
    \vspace{-10pt}
    \label{crystal}
\end{wrapfigure}translation of the unit cell $(\mathbf{A}, \mathbf{X})$  according to the periodic lattice matrix $\mathbf{L}$, as shown in Fig. \ref{crystal}.  
  The periodic lattice matrix $\mathbf{L}=\left[\mathbf{l}_1, \mathbf{l}_2, \mathbf{l}_3\right]$ is utilized to represent how a unit cell repeats itself in three directions. Specifically, considering a unit cell that includes $N$ atoms, $\mathbf{A}=\left[ \mathbf{a}_1,\mathbf{a}_2, \cdots, \mathbf{a}_N\right] \in \mathbb{R}^{N \times d_a}$ is the feature vector of atom type (i.e.,  chemical element), where $d_a$ is the dimension of atom feature vector. $\mathbf{X} = [\mathbf{x}_1, \mathbf{x}_2, \cdots, \mathbf{x}_N] \in \mathbb{R}^{N \times 3} $ denotes atom coordinates of each atom, where atom position $\mathbf{x}_{i} \in \mathbb{R}^3$ is represented  by Cartesian coordinates in 3D space. 
  Generally, given a crystal $\mathbf{C}$, its infinite periodic structure can be represented as utilizing a periodic lattice to repeat the unit cell in the 3D space as follows. 
\begin{equation}
   {\bf{A'}} = \{ {{\mathbf{a}'}_i}|{{{\mathbf{a}}'}_i} = {{\mathbf{a}}_i},1 \le i \le N\} 
   \setlength{\belowdisplayskip}{3pt}
\end{equation}
\begin{equation}
   {\bf{X'}} = \{ {\bf{x'}}_i|{{{\bf{x'}}}_i} = {{\bf{x}}_{{i}}} + {k_1}{{\bf{l}}_{{1}}} + {k_2}{{\bf{l}}_{{2}}} + {k_3}{{\bf{l}}_{{3}}},1 \le i \le N,{k_1},{k_2},{k_3} \in \mathbb{Z} \}
\end{equation}
where the integers $k_1$, $k_2$, and $k_3$ serve as translation vectors, enabling the unit cell to be replicated in three dimensions by means of the periodic lattice $\mathbf{L}$.

\vspace{-5pt}
\subsection{Quantum Chemical Constraints}
\vspace{-5pt}
Crystal structure analysis is subject to the fundamental constraints of quantum chemicals, including E(3) invariance (i.e., translation, rotation and reflection invariances) and periodic invariance. 

Specifically, E(3) invariance indicates that the unit cell's crystal structure remains unchanged when applying rotations and reflections to the coordinate matrix $\bf{X}$ and lattice matrix $\bf{L}$ simultaneously or applying translations to $\bf{X}$ alone, which can be formulated as follows.
\begin{equation}
    f({\bf{A}},{\bf{X,L}}) = f({\bf{A}},\mathcal{R}{\bf{X}} + {\bf{b,}}\mathcal{R}{\bf{L}})
\end{equation}
where $\mathcal{R} \in \mathbb{R}^{3 \times 3}$ is rotation and reflection transformations. $\bf{b} \in \mathbb{R}^3$ represents translation transformations in 3D space. Beyond the E(3) invariance, a crystal possesses its unique periodic invariance. As depicted in Fig. \ref{crystal}(c), the crystal is built up by repetitive translation of the unit cell. To elucidate the representation of an infinite crystal structure $(\mathbf{A'},\mathbf{X'})$, the principle of periodic invariance can be expressed via the subsequent formula.
\begin{equation}
    f({\bf{A}},{\bf{X,L}}) = f(g({\bf{A'}},{\bf{X'}},\alpha {\bf{L}},\beta ),\alpha {\bf{L}})
\end{equation}
where $g(\cdot)$ denotes utilizing a corner point $\beta \in \mathbb{R}^3$ and lattice matrix $\mathbf{L}$  to transform $(\mathbf{A'},\mathbf{X'})$ to $(\mathbf{A},\mathbf{X})$. $\alpha  \in \mathbb{N}_ + ^3$ denotes the scaling up of a repeating unit cell formed by periodic boundaries.

 \vspace{-5pt}
\section{Methodology}
 \vspace{-5pt}
\label{mmf}


\begin{figure}
    \centering
    \includegraphics[width=\columnwidth]{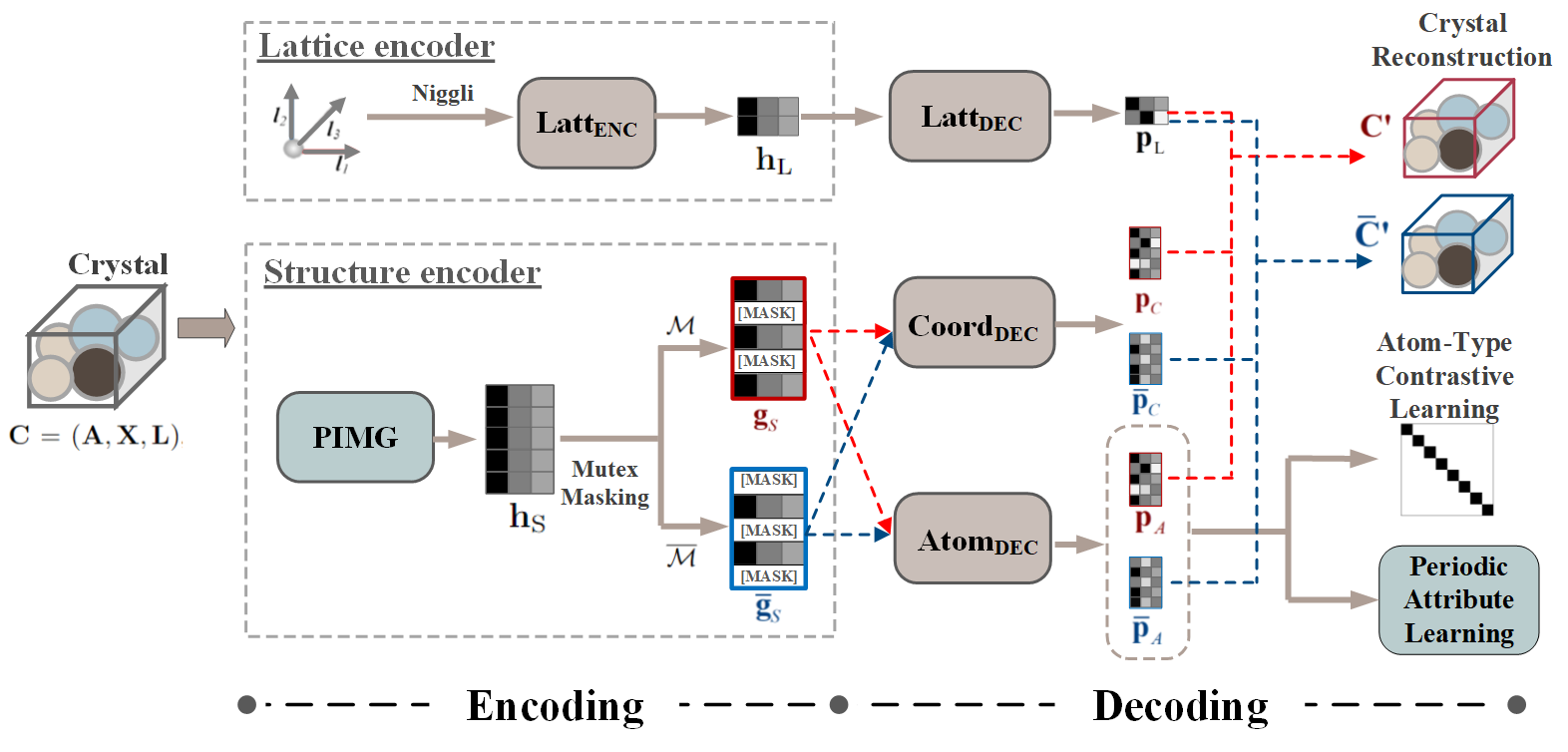}
    \caption{The pipeline of our mutex masked pre-training framework. During the encoding phase, MMPT consists of a lattice encoder and a structure encoder. The lattice encoder receives the lattice matrix and produces a lattice representation ${\bf{h}}_L$. The structure encoder first feeds the crystal material $\bf{C}$ into a periodic invariance multi-graph (PIMG) module and yields a structure representation ${\bf{h}}_S$.  Then,  mutex masking processes ${\bf{h}}_S$ to two  mutually exclusive views ${\bf{g}}_{S}$ and ${\overline{\bf{g}}}_{S}$.  During the decoding phase, the lattice decoder Latt$_{\rm{DEC}}$ decodes the lattice representation $\bf{h}_L$, and generates a reconstructed lattice representation ${\bf{p}}_L$. The coordinate decoder Coord$_{\rm{DEC}}$ processes the  mutex representations ${\bf{g}}_{S}$ and ${\overline{\bf{g}}}_{S}$, and produces reconstructed representations ${\bf{p}}_C$ and $\bar{\bf{p}}_C$. Similarly, the atom decoder Atom$_{\rm{DEC}}$ yields  reconstructed representations ${\bf{p}}_A$ and $\bar{\bf{p}}_A$. Following this, the decoding phase covers three tasks: crystal reconstruction, atom-type contrastive learning, and periodic attribute learning. }
    \label{fig:my_label}
 \vspace{-10pt}
\end{figure}

\subsection{Framework Overview}
 \vspace{-5pt}
We propose a Mutex Masked Pre-Training (MMPT) framework while ensuring periodic invariance for learning crystal representations, as depicted in Fig. \ref{fig:my_label}. 

During the encoding phase, our process involves two components: a lattice encoder and a structure encoder.  (1) The lattice encoder Latt$_{\rm{ENC}}$ encodes the lattice matrix $\mathbf{L}$ into a lattice representation $\mathbf{h}_{\rm L} \in \mathbb{R}^{d_l}$ that emphasizes the periodic information. Note that the lattice matrix $\mathbf{L}$ is not rotation invariant, and thus we pass it to Niggli Algorithm \cite{grosse2004numerically}, which can establish a set of conditions that determine a unique choice of basis vectors for a lattice.  (2) The structure encoder aims to encode  the crystal material $\mathbf{C}$ into a structure representation  ${{\bf{h}}_S} = [{\bf{h}}_S^1,{\bf{h}}_S^2,...,{\bf{h}}_S^N]\in N \times \mathbb{R}^{d_s}$. For capturing the structure representation while following E(3) and periodic invariances, we design a periodic invariance multi-graph (PIMG) module.   
 Then,  the structure representation ${\bf{h}}_S$ is masked randomly to produce two mutually exclusive views  ${\bf{g}}_{S}$ and ${\overline{\bf{g}}}_{S}$. 
 
During the decoding phase, the encoded representations are dispatched to various decoders.  The lattice representation $\bf{h}_L$ is sent to the lattice decoder Latt$_{\rm{DEC}}$, which yields a reconstructed lattice representation ${\bf{p}}_L$. The mutex representations ${\bf{g}}_{S}$ and ${\overline{\bf{g}}}_{S}$ are processed by the coordinate decoder Coord$_{\rm{DEC}}$, resulting in ${\bf{p}}_C$ and $\bar{\bf{p}}_C$. Concurrently, the mutex representations ${\bf{g}}_{S}$ and ${\overline{\bf{g}}}_{S}$ are also passed into the atom decoder Atom$_{\rm{DEC}}$, generating reconstructed representations ${\bf{p}}_A$ and $\bar{\bf{p}}_A$. Note that the atom decoder Atom$_{\rm DEC}$ and coordinate decoder Coord$_{\rm DEC}$ are set as Graph Isomorphism Network (GIN) \cite{xu2018powerful} to capture the relationships and dependencies between different atoms and substructures. Following this, the decoding process encompasses three tasks: crystal reconstruction, atom-type contrastive learning, and periodic attribute learning. (1) The crystal reconstruction task reconstructs two crystal materials from the reconstructed representations under two mutually exclusive structure views, i.e., $({\bf{h}}_L, {\bf{p}}_C, {\bf{p}}_A)$ and ${(\bf{h}}_L, {\bar{\bf{p}}}_C, \bar{\bf{p}}_A)$, as shown in the blue and red reconstructed material in Fig. \ref{fig:my_label}. (2) The atom-type contrastive learning is designed to emphasize atom-type representation learning, thereby encouraging the model to learn distinctive features. Emphasizing atom-type representation learning is inspired by the critical role of atom types in predicting crystal properties, as the chemical properties of atoms dictate their interactions within a crystal lattice  \cite{kittel2018introduction}.  (3) The periodic attribute learning (PAL) module is customized for reconstructing some periodic attributes by decoding the representations (i.e., $\mathbf{p}_A$ and $\mathbf{\bar p}_A$) generated by the atom decoder. This module can explicitly introduce periodic attributes while adhering to crystal quantum chemistry principles,  including periodic invariance.
 
 \vspace{-5pt}
 \subsection{Mutex Mask Strategy}
  \vspace{-5pt}
 






\noindent{\textbf{Motivation.}} Inspired by the successful masked language modeling technique used in BERT\cite{devlin2018bert}, we introduce masking techniques into our pre-training framework. This is because the relationships among atoms in a crystal structure can be analogous to the contextual relationships among words in a sentence. However, random masking representation might focus on a specific pattern or feature in the data and ignore others. That is to say, if the masked parts of the data consistently relate to a specific feature or pattern, the model may learn to over-rely on this feature.  Thus, we design a mutex masking strategy that has two mutually exclusive views for capturing informative and diverse representations. This is due to the model being exposed to different perspectives of the data, allowing it to capture a richer set of features. Therefore, the mutex masking strategy can lead the model to handle a wider variety of patterns and features under two mutually exclusive views.

\noindent{\textbf{Mutex Masking.}} 
\label{mutex_mask}
The mask strategy corrupts the encoding representing  ${\bf{h}}_S$ on purpose to enforce the model to learn the representations that take into account surrounding atoms and their relationships, rather than learning individual atom representations. We first give a notation of mutex masks.
Let's consider a full feature set with $N$ indices, i.e., $\{1,..., N\}$. If one mask randomly selects part indices with a uniform distribution $\cal{M}$, then the mask processing replaces their subset of feature as [MASK].
Another mask replaces the complementary set of indices, denoted as $\overline {\cal{M}} = $\{1,..., N\}$ \setminus {\cal{M}}$. We refer to this pair of masks (i.e., $\cal{M}$ and $\overline {\cal{M}}$), which operate on disjoint sets of indices, as \emph{mutex masks}.

Specifically, for the structure representation  ${{\bf{h}}_S} = [{\bf{h}}_S^1,{\bf{h}}_S^2,...,{\bf{h}}_S^N]$, we randomly select a subset of atoms $\mathcal{M}$ and replace their subset of features with the $[{\rm MASK}] \in \mathbb{R}^{d_s}$ token. The masking  process can be formulated as follows. 
\begin{equation}
{\bf{g}}_S^i = \left\{ 
\begin{array}{l}
{\bf{h}}_S^i, i \notin {\cal M}\\

[{\rm{MASK}}],i \in {\cal M}
\end{array} \right.
 \setlength{\belowdisplayskip}{6pt}
\end{equation}
where $\mathbf{g}_{\rm S}=[\mathbf{g}_{\rm S}^{1},\mathbf{g}_{\rm S}^{2},...,\mathbf{g}_{\rm S}^{N}] \in N \times \mathbb{R}^{d_s}$ is the structure feature after masking processing, and $[{\rm MASK}]$ is a fixed token that replaces the features of the masked nodes. 
 We not only use a subset of ${\cal M}$ to mask the structure representation ${{\bf{h}}}_S$ as ${{\bf{g}}}_S$, but also we use its corresponding mutex subset ${\overline {\cal M} }$ to mask the structure representation ${{\bf{h}}}_S$ producing the ${\overline{\bf{g}}}_S$. The masking process is critical in our framework as it forces the model to learn representations according to the structure relationship between two complementary sets of atoms rather than solely dependent on the features  of each atom. Under these two  mutually exclusive views, we design a crystal reconstruction task and an atom-type contrastive learning task for our pre-training framework. We detail these two tasks as follows.

\noindent{\textbf{Crystal Reconstruction.}} 

After passing the mask representation to the decoder, we reconstruct the three core components $\mathbf{A}$, $\mathbf{X}$ and $\mathbf{L}$ of the crystal by optimizing reconstruction loss $\cal{L}_{REC}$. Note that we only calculate the loss related to the masked index of atoms. The reconstruction loss includes (1) Atom types $\mathbf{A}$ are predicted by minimizing the cross entropy ${\cal{L}}_{\bf{A}}$ between the ground truth atom types and predicted atom types. (2) Considering the crystal should follow rotation invariant, we calculate the distance of each atomic coordinate from the center coordinate, and use this distance as the prediction target rather than using the coordinate position directly. Hence, the target loss ${\cal{L}}_{\bf{X}}$ for reconstructing coordinates is based on distances. (3) The lattice ${\bf{L}} \in \mathbb{R}^{3 \times 3}$ is simplified to six unique, rotation-invariant parameters using the Niggli algorithm. These include the lengths of the three lattice vectors and the angles between them. We optimize the prediction of the lattice through an ${\cal{L}}_{\bf{L}}$ loss function. More details can be found in Appendix \ref{lossf}.



\noindent{\textbf{Atom-Type Contrative Learning.}} Atom type is a critical factor in predicting crystal properties because the chemical properties of atoms determine how they interact with one another in a crystal  lattice \cite{kittel2018introduction}. Different atom types have distinct electron configurations and bonding characteristics, which can significantly influence the crystal's properties.  Therefore, besides reconstructing the masked representations of crystal material, we also introduce constraints for atom types.
Considering the Barlow Twins approach is a specific type of contrastive learning that can learn useful representations of data while reducing redundancy between input vectors, we utilize it to further constraint atom-type representation learning. 
Our objective is to foster the learning of informative and complementary representations by designing mutex masks to promote the generation of such representations.
Considering the importance of atom type for property prediction, we introduce constraints for ${{\bf{p}}}_A^i$ and ${\overline{\bf{p}}}_A^i$  by calculating the Barlow Twins loss. We measure the cross-correlation matrix between two vectors ${{\bf{p}}}_A^i$ and ${\overline{\bf{p}}}_A^i$  and make it as close to the identity matrix as possible. More details of Barlow Twins loss can be found in Appendix \ref{lossf}.

 \vspace{-8pt}

\subsection{Periodic-Specific Modules}
 \vspace{-8pt}
\label{csm}
To explicitly encode periodic-related attributes, we introduce periodic-specific modules, including a periodic invariance multi-graph (PIMG) module and a periodic attribute learning (PAL) module. 

\noindent{\textbf{Periodic Invarince Multi-Graph Module.}}
\begin{figure}
 \vspace{-5pt}
    \centering    \includegraphics[width=0.95\columnwidth]{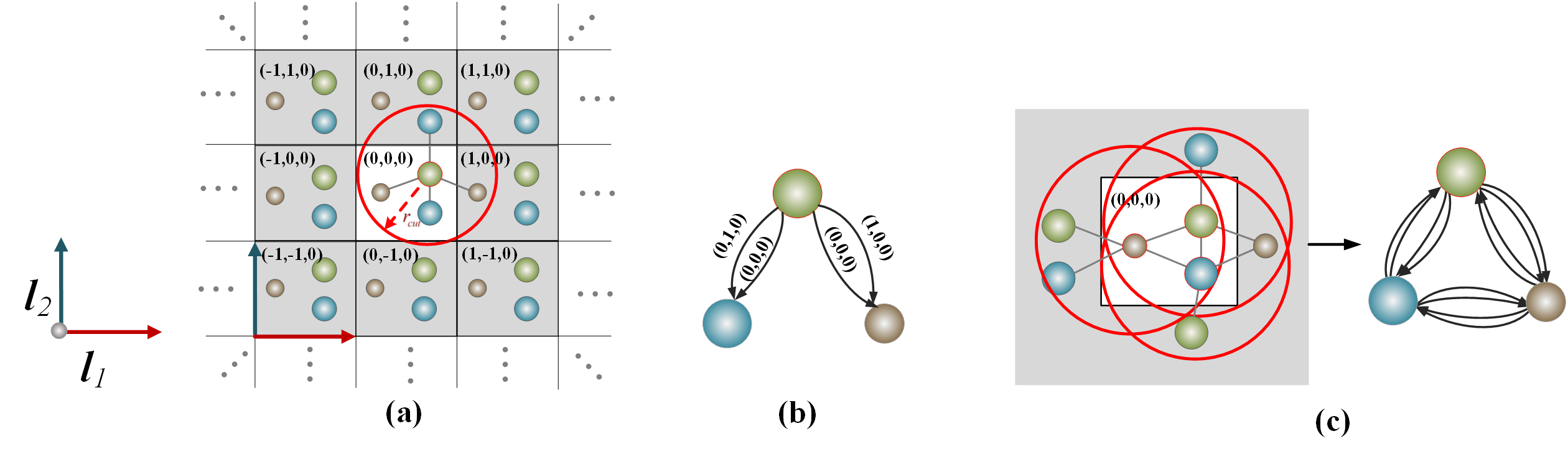}
    \caption{Multi-graph construction for a crystal with 3 atoms. Shown in a 2D space for easy illustration, by setting $k_3=0$ and only repeating the unit cell along the ${\bf{l}}_1$ and ${\bf{l}}_2$ directions.   (a) Illustration of the multi-graph construction for the green atom. (b) Illustration of the multi-edges for the green atom with $(k_1, k_2, k_3)$ labeling on top of the edges. (c) Illustration of the multi-graph construction for the center unit cell, which includes three atoms. Each red circle signifies the cutoff circle with the neighbors of each atom in the unit cell. }
    \label{fig:mg_r}
    \vspace{-15pt}
\end{figure}
Crystal materials can be denoted as a multi-graph $\mathbf{G}={\{\mathbf{A}, \mathbf{E}\}}$ to learn the structure of the crystal material, as shown in Fig.\ref{fig:mg_r} (a). $\mathbf{A}=\left[ \mathbf{a}_1,\mathbf{a}_2, \cdots, \mathbf{a}_N\right] \in \mathbb{R}^{N \times d_a}$ is the set of atom nodes in the crystal, which is the feature vector of atom type, as defined in Sec. \ref{cryrep}. $\mathbf{E} = \{e_{ij, (k_1, k_2, k_3)} | i, j \in \{1, ..., N\}, k_1, k_2, k_3 \in \mathbb{Z}\}$ is the set of edges representing relevant atom pairs in crystal. The edge $e_{ij, (k_1, k_2, k_3)}$ denotes a directed edge from node $i$ in the original unit cell to node $j$ in the cell translated by $k_1\mathbf{l}_1 + k_2\mathbf{l}_2 + k_3\mathbf{l}_3$, where $\mathbf{l}_1, \mathbf{l}_2$, and $\mathbf{l}_3$ are the lattice vectors of the crystal. As shown in Fig. \ref{fig:mg_r} (b), the values of $(k_1, k_2, k_3)$  labeling on top of the edges show the translation operation. For example, (0,0,0) indicates that the edge connection resides within the same unit cell, while (0,1,0) signifies that the connection exists in a neighboring unit by translating the vector $\mathbf{l}_2$.
To construct edges for relevant atom pairs while following periodic invariance,  k-nearest neighbor (KNN) approaches with cutoff distances $r_{cut}$ are often utilized to determine its edges in the graph, as shown in Fig. \ref{fig:mg_r} (c). To ensure the multi-graph follows the periodic invariance by constructing edges between atom node $i$ and atom node $j$ by calculating the Euclidean distance: $\{ {d_{ij}}|{d_{ij}} = {\left\| {{{\bf{x}}_i} - {{\bf{x}}_j} + {k_1}{{\bf{l}}_1} + {k_2}{{\bf{l}}_2} + {k_3}{{\bf{l}}_3}} \right\|_2},{k_1},{k_2},{k_3} \in \mathbb{Z},{d_{ij}} \le r_{cut}, 1 \le j \le n \} $, where $n$ is the neighbor number. The pairwise Euclidean distances are invariant to the shifts of periodic boundaries, which is proved in \cite{yanperiodic}. 

To capture complex and subtle structural patterns adaptively, we design a multi-graph attention mechanism. 
First, a shared linear transformation, parameterized by a weight matrix $\bf{W} \in \mathbb{R}^{{d_a}' \times d_a}$, is applied to the initial feature vector of each atom node $\mathbf{a}_i$. Then, we perform attention on the atom nodes, and the reweighted atom node ${{\hat{\bf{a}}}_i}$ is calculated by:
\begin{equation}
\setlength{\abovedisplayskip}{6pt}
    {r_{ij}} = {f_a}({\bf{Wa}}_i,{\bf{Wa}}_j),{\varepsilon_{ij}} = \frac{{\exp ({\rm{LeakyReLU(}}{r_{ij}}{\rm{)}})}}{{\sum\limits_{k \in {\cal{N}}_i} {\exp ({\rm{LeakyReLU(}}{r_{ik}}{\rm{)}})} }}, {{\hat{\bf{a}}}_i} = FC(\sum\limits_{j \in {\cal{N}}_i} {{\varepsilon _{ij}} \cdot } {\rm{ }}{\bf{a}}_i)
    \setlength{\belowdisplayskip}{6pt}
\end{equation}
where ${f_a}$  is a single-layer feed-forward neural network. ${\varepsilon_{ij}}$ is a correlation coefficient that is normalized by the softmax function. The correlation coefficient semantically indicates the importance of the neighbor node $j$ to the target node $i$. $LeakyReLU(\cdot)$ is the activation function, and  ${\cal{N}}_i$ represents the neighbor nodes set of node $i$. $FC$ is a fully-connected layer. 

Finally, we sent the multi-graph and the reweighted atom representation $\mathbf{\hat{A}}=\left[ \mathbf{\hat{a}}_1,\mathbf{\hat{a}}_2, \cdots, \mathbf{\hat{a}}_N\right]$ into an E(3) invariance graph neural network to learn the crystal structure representations from message passing. For this purpose, we employ DimeNet++ \cite{gasteiger2020fast}, which satisfies the E(3) invariance requirement, as the graph neural network.




\noindent{\textbf{{Periodic Attribute Learning Module.}}
We customize a periodic attribute learning module for the self-supervised learning framework. This module focuses on predicting three types of periodic attributes related to periodicity from the  representations, which can introduce learning constraints to enforce model learning more robust representations.
For each edge, we concatenate the $i$-th node of crystal $\bf{C}$ with the representations of its neighboring atom nodes and obtain $ {\bf{p}}_A^{_{ij}} = [{\bf{p}}_A^i||{\bf{p}}_A^j]$ and ${\bf{\bar p}}_A^{_{ij}} = [{\bf{\bar p}}_A^i||{\bf{\bar p}}_A^j]$. These mutex  representations (i.e., $ {\bf{p}}_A^{_{ij}}$ and ${\bf{\bar p}}_A^{_{ij}}$) are then fed into three types of attribute learners (i.e., multilayer perceptron) for predicting three periodic-specific attributes: discrete direction, unit cell position, and distance between nodes. Then, we optimize the attribute prediction by calculating $\cal{L}_{CAA}$ loss, which includes ${\cal{L}}_{Die}$, ${\cal{L}}_{Unit}$  and ${\cal{L}}_{Dis}$ losses (cf. Appendix \ref{lossf}).

More specifically, given a node with coordination $\mathbf{x}_i$, the coordination of its corresponding neighbor $j$ can be represented by  $ {{{\bf{x}}}_j} = {{\bf{x'}}_{{j}}} + {k_1}{{\bf{l}}_{{1}}} + {k_2}{{\bf{l}}_{{2}}} + {k_3}{{\bf{l}}_{{3}}}$. The crystal attributes between edge $\mathbf{x}_i$ and  $ {{{\bf{x}}}_j}$ are detailed as follows.
(1) \emph{Discrete Direction}. We determine the directions between each atom and its corresponding neighboring atoms. We streamline the process by discretizing the directions of each atom pair into 27 distinct directions in 3D space.  If atom pairs are in the same unit~(i.e., $k_1, k_2, k_3=0$ ), we regard this situation as the same direction. Thus, the directions can be discretized  by the arrangement and combination of the $k_1, k_2, k_3$ after the modulo operation, i.e., $\frac{{{k_1}}}{{|{k_1}|}},\frac{{{k_2}}}{{|{k_2}|}},\frac{{{k_3}}}{{|{k_3}|}} \subseteq \{ 1,0, - 1\} $.   To better illustrate this concept, we show it in 2D space, thereby limiting us to showing only 9 directions, as demonstrated in Fig. \ref{fig:crystal} (a). (2) \emph{Unit Cell Position}. The unit cell position attribute is used to determine whether two connected atomic pairs belong to the same unit cell, as shown in Fig. \ref{fig:crystal} (b). In detail, for each pair of atoms, if both atoms are located within the same unit cell, they can be considered as having an internal connection, i.e.,  $k_1, k_2, k_3 = 0$. Otherwise,  the atom pairs are considered to have external connections, i.e.,  $\{(k_1, k_2, k_3) \in \mathbb{Z}^3 \mid (k_1, k_2, k_3) \neq (0, 0, 0)\}$. (3) \emph{Edge Distance}. 
As shown in  Fig. \ref{fig:crystal} (c), we predict the distance $d_{ij}=||\mathbf{x}_i- {\mathbf{x}}_j||$ between node ${\bf{x}}_i$ and node ${\bf{x}}_j$ to enhance our MMPT framework.
\begin{figure}[!h]
    \centering
     
    \includegraphics[width=0.75\columnwidth]{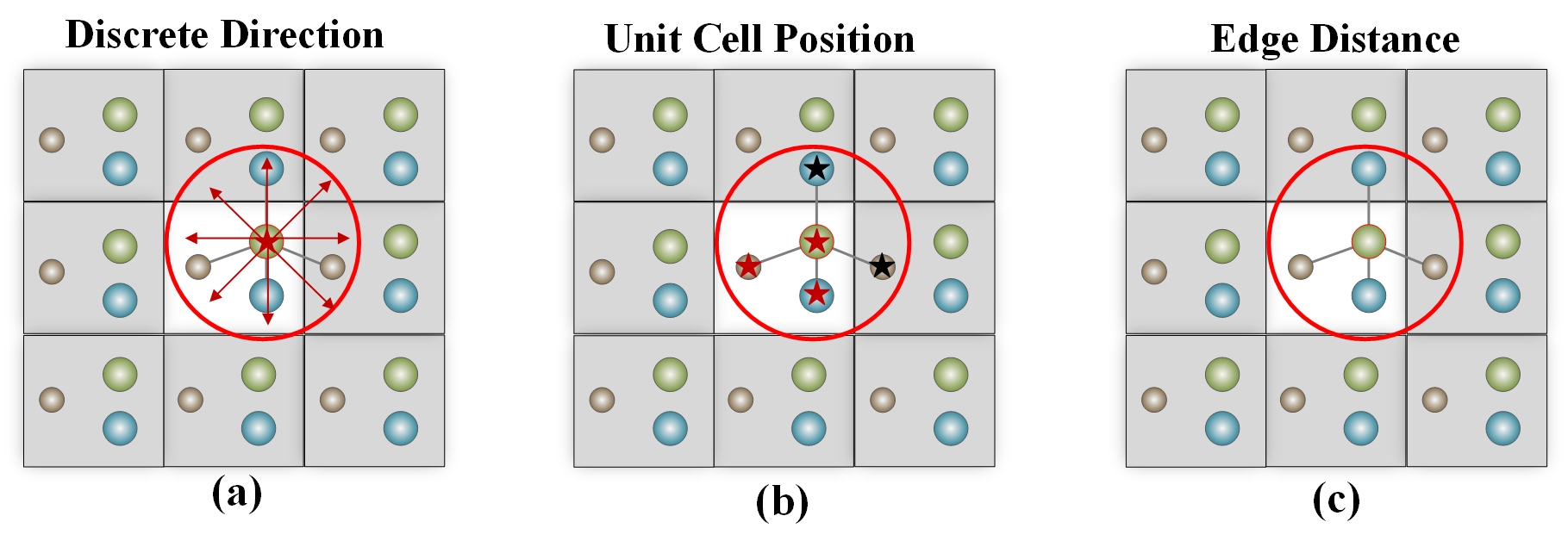}
    \caption{The illustration of the periodic attribute learning module. We utilize the green atom in the center unit cell as a target. (a) Illustration of discrete direction from green atom to its neighbors. (2) Illustration of whether the green  atom and its neighbors are in the same unit cell. (3) Illustration of the distance between green atom and its neighbors.  }
    \label{fig:crystal}
     \vspace{-10pt}
\end{figure}

We focus on predicting these three types of crystal attributes related to periodicity to enforce model learning more robust representations.

 \vspace{-5pt}
\section{Experiments}

 \vspace{-8pt}
\subsection{Implementation Details}
 \vspace{-8pt}
To pre-train the framework, we use a subset \cite{choudhary2020joint} of the Open Quantum Materials Database (OQMD) \cite{kirklin2015open, saal2013high} with 817139 material structures. The OQMD \cite{kirklin2015open, saal2013high} offers a substantial amount of unlabeled data that proves sufficient for the purpose of pre-training.
 To evaluate the performance of the pre-training framework, we subjected them to rigorous testing on a set of 8 challenging benchmark datasets,  including datasets with limited data samples. To provide a summary of these 8 datasets, Table \ref{dataset} has been included. More detailed descriptions of these datasets are available in Appendix \ref{data_append}. For downstream applications, the structure encoder is applied without any masking strategy, and the decoding stage is only used during the self-supervised training stage rather than for the downstream application. More details are shown in 
Appendix \ref{MMPT_config}.

\begin{wraptable}{r}{0.6\columnwidth}
 \vspace{-10pt}
\caption{Summary of eight different datasets.}
\smallskip\resizebox{0.6\columnwidth}{!}{
\begin{tabular}{c|c|c|c}
\toprule[1.4pt]
  Dataset         & \#Crystals & Property           & Unit         \\ \hline
 
JDFT2D\cite{choudhary2017high,dunn2020benchmarking}       & 636        & Exfoliation Energy & meV/atom \\ \hline
 Dielectric\cite{jain2013materials,dunn2020benchmarking}     & 4,764      & Refractive Index   & Unitless     \\ \hline
Mp\_Shear \cite{chen2019graph}      & 5,449       & Shear Modulus      & log$_{10}$VRH     \\ \hline
 Mp\_Bulk \cite{chen2019graph}       & 5,450       & Bulk Modulus       & log$_{10}$VRH     \\ \hline
KVRH \cite{de2015charting,dunn2020benchmarking}           & 10,987     & Bulk Modulus       & log$_{10}$VRH     \\ \hline
 Jarvis\_gap \cite{choudhary2021atomistic} & 18,171     & Band gap           & eV           \\ \hline
 Jarvis\_Ehull \cite{choudhary2021atomistic}  & 55,370     & Ehull              & eV           \\ \hline
 Mp\_gap  \cite{chen2019graph}   & 69,239     & Band gap           & eV           \\ 
\toprule[1.4pt]
\end{tabular}}
\vspace{-8pt}
\label{dataset}
\end{wraptable}
The proposed method is conducted on a device with NVIDIA TITAN RTX 24GB GPU and implemented with Pytorch. The code will be made publicly available upon acceptance. To compare the performance of each model, we utilize mean absolute error (MAE) for property prediction.  More detailed implementations of the hyperparameters are available in Appendix \ref{MMPT_config}. To evaluate the effectiveness of our framework, we compare our proposed framework with both supervised and self-supervised based baselines, including CGCNN \cite{xie2018crystal}, SchNet \cite{schutt2017schnet}, MEGNet \cite{chen2019graph}, ALIGNN \cite{choudhary2021atomistic}, Matformer \cite{yanperiodic}, and Crystal Twins \cite{magar2022crystal}. The Crystal Twins is a self-supervised-based method, while others are supervised-based methods.  In our experiments, we also pre-train the Crystal Twins model using data from the OQMD dataset. The baselines are detailed in Appendix \ref{baseline}.

\vspace{-8pt}

\subsection{Experimental Results}
\vspace{-5pt}
Table \ref{result} presents the performance compared with both supervised-based methods and self-supervised-based methods on eight datasets. The decimal precision reported in the experiments follows previous works \cite{yanperiodic,dunn2020benchmarking}. The best and the second best results are in bold and underlined, respectively. We analyze the experimental results in detail.
(1) Overall, MMPT achieves the best performance accuracy compared with supervised-based methods and self-supervised-based methods. In addition, the improvement for our framework is more significant than that on the small data sets. This suggests that MMPT is effective in alleviating the limited number labeled. (2) Our proposed MMPT performs better than the self-supervised Crystal Twins method. This method does not take full advantage of the E(3) invariant and period invariant of the crystal. Therefore, fully considering the E(3) invariance and period invariance of the crystal is the key to improving the performance.
\begin{table}[!h]
\caption{Comparison between MMPT and other baselines in terms of test MAE on eight datasets.}
\renewcommand{\arraystretch}{1.1}
\centering
 \smallskip\resizebox{\columnwidth}{!}{
\begin{tabular}{c|c|c|c|c|c|c|c|c}
\toprule[1.6pt]
Method                                                                                & JDFT2D                     & Dielectric      & Mp\_Shear      & Mp\_bulk       & KVRH            & Jarvis\_gap & Jarvis\_ehull & Mp\_gap \\ \hline
\#   Crystals                                                                         & 636                        & 4764            & 5449           & 5450           & 10987           & 18171           & 55370         & 69239       \\ \hline
\begin{tabular}[c]{@{}c@{}}SchNet\\ (2017,NeurIPS)\end{tabular}                       &  \underline{42.6637}                   &  \underline{0.3277}          & 0.099          & 0.066          & 0.0590          & 0.43            & 0.140          & 0.345       \\ \hline
\begin{tabular}[c]{@{}c@{}}CGCNN\\  (2018, Phys. Rev. Lett.)\end{tabular}             & 49.2440                  & 0.5988          & 0.077          & 0.047          & 0.0712          & 0.41            & 0.170          & 0.292       \\ \hline
\begin{tabular}[c]{@{}c@{}}MEGNet \\ (2019, Chem. Mater.)\end{tabular}                & 54.1719                 & 0.3391          & 0.099          & 0.060          & 0.0668          & 0.34            & 0.084         & 0.307       \\ \hline
\begin{tabular}[c]{@{}c@{}}ALIGNN\\ (2021, NPJ Comput. Mater.)\end{tabular}           & 43.4244                   & 0.3449          & 0.078          & 0.051          &  \underline{0.0568}          & 0.31            & 0.076         & 0.218       \\ \hline
\begin{tabular}[c]{@{}c@{}}Matformer\\ (2022,NeurIPS)\end{tabular}                    &  47.6964                &   0.6817        &  \underline{0.073}          &  \underline{0.043}          &   0.0620        &  \underline{0.30}            &  \underline{0.064}         &  \underline{0.211}       \\ \hline

\begin{tabular}[c]{@{}c@{}}Crystal Twins \\ (2022,   NPJ Comput. Mater.)\end{tabular} &  44.3536           &            0.4276      &        0.082        &    0.050            &     0.0665           &     0.39            &      0.140           &   0.291                         \\ \hline
MMPT                                                                                   & \textbf{38.2133} &  \textbf{0.3240} & \textbf{0.072} & \textbf{0.038} & \textbf{0.0567} & \textbf{0.26}   &   \textbf{0.061}            &    \textbf{ 0.204}        \\ \toprule[1.6pt]          \end{tabular}}
\label{result}
\vspace{-8pt}
\end{table}
\vspace{-5pt}
\subsection{Ablation Studies}
 \vspace{-5pt}
To explore the effectiveness of each component of MMPT, we design ablation studies to compare our MMPT with different variants. (1) MMPT-NO: The parameters of the pre-training framework are not shared with the downstream tasks. Instead, these tasks are trained from scratch. (2) MMPT-RE: MMPT-RE removes the crystal reconstruction architecture in MMPT.  (3) MMPT-MUTEX: It is a variant of MMPT, which only use only one mask rather than  mutex masks. (4) MMPT-CL: MMPT-CL removes
the atom-type contrastive learning in MMPT framework. (5) MMPT-PIMG: MMPT-PIMG removes the periodic invariance multi-graph module in MMPT framework. (6) MMPT-PAL: MMPT-PAL removes the  periodic attribute learning module in MMPT framework. 
The ablation studies are conducted on three datasets, including Mp\_bulk, Jarvis\_gap and Mp\_gap. 

As illustrated in Fig. \ref{fig:ab}, the experimental results reveal the following observations: (1) The best performance is achieved by MMPT, indicating that each component contributes effectively to crystal property prediction. (2) The removal of pre-training leads to a significant decline, which proves the importance of a pre-training framework. 

\begin{figure}
    \centering
    \includegraphics[width=\columnwidth]{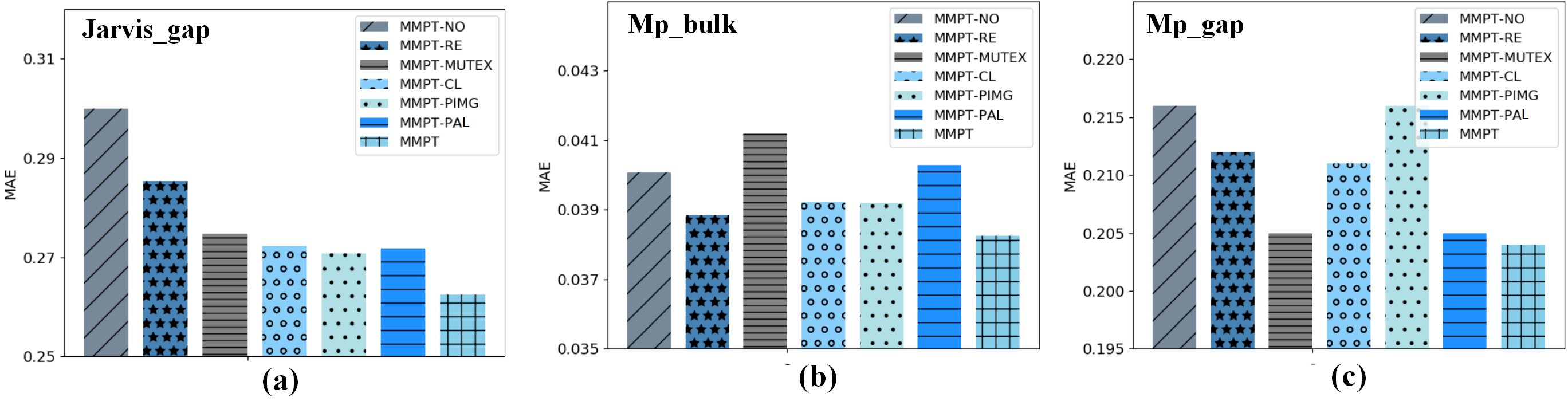}
    \caption{Ablation studies of MMPT on (a) Jarvis\_gap, (b) Mp\_Buk and (c) Mp\_gap Datasets.}
    \label{fig:ab}

\end{figure}

Moreover, we compared the performance of training downstream tasks with our pre-training framework versus without a pre-training framework on three datasets: Jarvis\_gap, MP\_Bulk and Mp\_gap. We assessed the performance in both cases by using 10\%, 30\%, 50\%, 70\%, 90\%, and 100\% of the samples for training during the fine-tuning process on these datasets. For example, as depicted in Fig. \ref{fig:percent}(a), when the data ratio of the MP\_Bulk dataset decreases from 100\% to 10\%, the MAE loss gap between the two methods continues to increase from 0.03 to 0.08. With pre-training, we can achieve the same performance as with non-pre-trained situations while using fewer labeled training data. These results demonstrate that employing our pre-trained model offers more advantages than not using it, especially when the amount of labeled data is limited.

\begin{figure}[!h]
    \centering
    \includegraphics[width=\columnwidth]{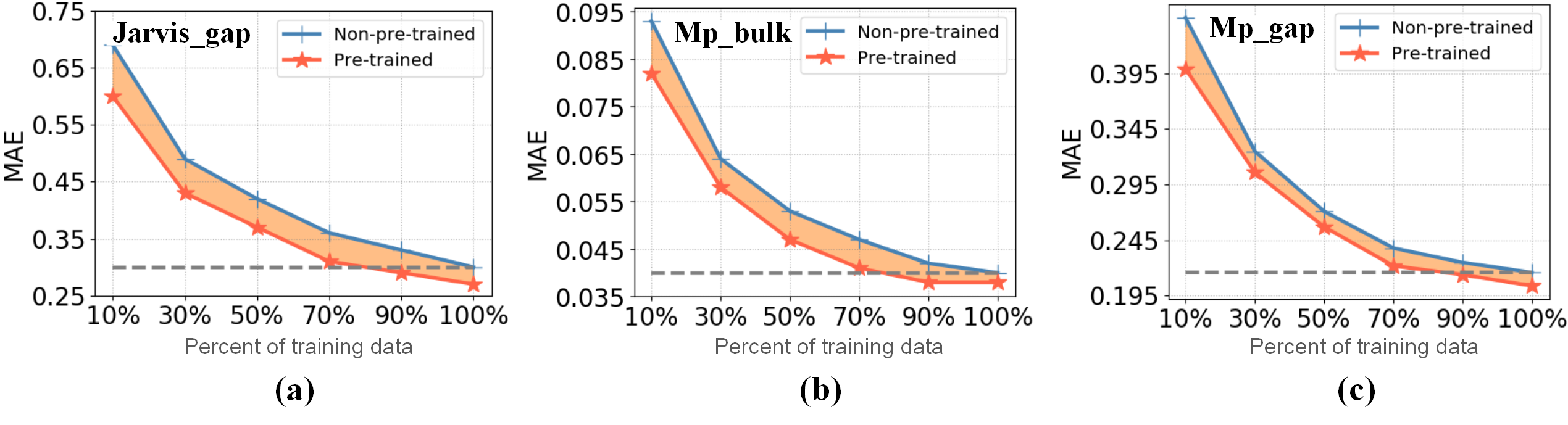}
    \caption{Effects of Pre-training (a) Jarvis\_gap, (b) Mp\_Buk and (c) Mp\_gap Datasets.}
    \label{fig:percent}
 \vspace{-15pt}
\end{figure}

\section{Conclusion}
In this paper, we proposed a mutex masked pre-training (MMPT) framework while ensuring periodic invariance for learning crystal representations with self-supervision. The MMPT framework has been developed to tackle the challenge of limited labeled data, concurrently satisfying and fully leveraging the distinctive periodic invariance inherent to crystals. The mutex mask strategy can force the model to learn the representations under two mutually exclusive views that take into account surrounding atoms and their relationships, rather than learning individual atom representations. The periodic invariance multi-graph module and periodic attribute learning further enhance the performance of the MMPT framework. The experiment proves that the mutex masked pre-training can effectively improve the prediction performance for crystal properties.  In the future, it is of interest to further explore the chemical characteristics of each property
 to find the features that are more helpful for crystal property prediction.
\bibliography{neurips}
\appendix
\newpage
\section{Appendix}
\subsection{Dataset Descriptions}
\label{data_append}
In our paper, we exploit a dataset for the pre-training phase and deploy eight datasets for various downstream tasks. The  objective of these eight downstream tasks is to evaluate the performance of our proposed pre-training framework. First, the dataset used for pre-training tasks is described below:

\textbf{OQMD \cite{kirklin2015open, saal2013high, choudhary2020joint}}: This dataset includes the data from the Open Quantum Materials Database (OQMD) \cite{kirklin2015open, saal2013high} and is obtained by JARVIS-Tools \cite{choudhary2020joint}, an open-access software package for atomistic data-driven materials computation. It has 817636 material structures. To ensure the quality of the dataset, we filtered out the materials with extreme formation energy that is either above 4.0 or below -5.0. We also filter out a crystal structure \footnote{https://oqmd.org/materials/entry/1339536}  that cannot be successfully loaded. We use a cleaned dataset with 817139 material structures for the pre-training task. Each structure is saved as a CIF file and contains features of atom type, atom coordinate, and lattice.

Then, we detail the eight datasets for downstream tasks.
Among the datasets of downstream tasks, JDFT2D \cite{choudhary2017high, dunn2020benchmarking}, Dielectric \cite{jain2013materials, dunn2020benchmarking}, and KVRH \cite{jain2013materials, de2015charting, dunn2020benchmarking} are taken from the MatBench \cite{dunn2020benchmarking} task suite. The datasets of these tasks are divided into five folds for training, validation, and testing. Thus, for these three tasks, we report the average MAE on the five folds. For all the other tasks, we follow the settings in Matformer \cite{yanperiodic}. The datasets used for downstream tasks are described in detail below:


\textbf{JDFT2D \cite{choudhary2017high,dunn2020benchmarking}}: This dataset aims to predict exfoliation energies based on the crystal structure. The calculations are performed using the OptB88vdW and TBmBJ functionals. It consists of 636 samples.

\textbf{Dielectric \cite{jain2013materials,dunn2020benchmarking}}: The objective of this dataset is to predict the refractive index from the structure. Entries with formation energy (or energy above the convex hull) higher than 150meV, refractive indices less than 1, and noble gases are removed. The dataset comprises 4764 samples.

 \textbf{Mp\_Shear \cite{chen2019graph, jain2013materials}}: This dataset focuses on predicting the DFT log$_{10}$VRH-average shear modulus using the crystal structure. 
We have adopted identical training, validation, and testing sets as those used by GATGNN \cite{louis2020graph}, comprising 4664, 393, and 393 crystals, respectively. Note that one sample from the validation set was omitted due to a negative GPa value in the shear modulus.

\textbf{Mp\_Bulk \cite{chen2019graph, jain2013materials}}: For the Bulk Moduli dataset, the goal is to predict the DFT log$_{10}$VRH-average bulk modulus based on the structure. We followed the training, validation, and testing sets of GATGNN \cite{louis2020graph}, which include 4664, 393, and 393 crystals, respectively.

\textbf{KVRH \cite{jain2013materials, de2015charting,dunn2020benchmarking}}: The KVRH dataset aims to predict the DFT log$_{10}$ VRH-average bulk modulus from the structure. It excludes entries with formation energy (or energy above the convex hull) exceeding 150meV, negative G\_Voigt, G\_Reuss, G\_VRH, K\_Voigt, K\_Reuss, or K\_VRH values, and entries that fail G\_Reuss <= G\_VRH <= G\_Voigt or K\_Reuss <= K\_VRH <= K\_Voigt conditions. Noble gases are also excluded. The dataset consists of 10987 samples.

\textbf{Jarvis\_gap \cite{choudhary2021atomistic}}: This dataset focuses on predicting the band gap, calculated using the TBmBJ functionals. The training, validation, and testing sets contain 14537, 1817, and 1817 samples, respectively.

\textbf{Jarvis\_Ehull \cite{choudhary2021atomistic}}: The goal of this dataset is to predict the ehull. The training, validation, and testing sets include 44296, 5537, and 5537 samples, respectively.

 \textbf{Mp\_gap \cite{chen2019graph, jain2013materials}}: This dataset aims to predict the band gap. We followed the same training, validation, and testing sets as Matformer \cite{yanperiodic}, consisting of 60000, 5000, and 4239 crystals, respectively.

\subsection{MMPT  Configurations}
\label{MMPT_config}

We detail our MMPT configurations in the pre-training stage as follows. 
During the training process, the learning rate is set as 1e-5. The random seed is set as 123. The model parameters are optimized by Adam. We train our model for 50 epochs and select the model with the lowest loss.

\noindent \textbf{Atom type feature}
The feature vector of atom type $\mathbf{A}=\left[ \mathbf{a}_1,\mathbf{a}_2, \cdots, \mathbf{a}_N\right] \in \mathbb{R}^{N \times d_a}$ ( i.e.,  chemical element ) is obtained from embedding the one-hot feature $\mathbf{V}=\left[ {\bf{v}}_1,{\bf{v}}_2, \cdots, {\bf{v}}_N\right] \in \mathbb{R}^{N \times 119}$ of atom type. The dimension of $d_a$ is 128. In our work, the total number of atom-type classes is determined based on the number of chemical elements and is therefore set as 119.

\noindent \textbf{Lattice Encoder}
The lattice encoder Latt$_{\rm{ENC}}$ utilizes the Tranformer model, which we utilize the Block in timm.models.vision\_transformer. The embedding dimension is 8, the number of heads is 2, and mlp\_ratio is 4.
The dimension of $d_l$ is 16.

\noindent \textbf{Structure Encoder}
The structure encoder aims to encode the crystal material $\mathbf{C}$ into a structure representation ${\bf{h}}_S$. 
The periodic invariance multi-graph (PIMG) module first passes the atom feature vector into a multi-graph attention mechanism and then utilizes the DimeNet++ model to learn the structure representation. The implementation of the attention mechanism is referred to \footnote{https://github.com/gordicaleksa/pytorch-GAT/}. The number of heads is set as 1. Simultaneously, both the number of input features and output features are set to 128. The configuration of DimeNet++ is detailed as follows. The number of building blocks is set as 4, the size of embedding in the interaction block is 64, the size of basis embedding in the output block is 256, the number of spherical harmonics is 7, the number of radial basis functions is 6, the  cutoff distance $r_{cut}$  for interatomic interactions is set as 8.0, the maximum number of neighbors $n$ to  collect for each node is 12,  the shape of the smooth cutoff is set as 5, the number of residual layers in the interaction blocks before the skip connection is 1, the number of residual layers in the interaction blocks after the skip connection is 2, the number of linear layers for the output blocks is 3, and the activation function is "swish". The dimension of $d_s$ is 64. $|\cal{M}|$ is set as $N/2$.

\noindent \textbf{Lattice Decoder}
The lattice decoder utilizes five layers of fully connected linear with  RELU activation function to decoder lattice representation ${\bf{h}}_{L}$. 

\noindent \textbf{Coordinate Decoder}
The coordinate decoder utilizes a graph isomorphism network, the dimensionality of embeddings for nodes and edges is 64, and the aggregation method is ``add.''

\noindent \textbf{Atom Decoder}
The atom decoder utilizes a graph isomorphism network, the dimensionality of embeddings for nodes and edges is also set as 64, and the aggregation method is ``add.'' Then it is followed by a five-layer fully connected linear with RELU activation function to further decode the atom types.

In the fine-tuning stage, we share the weights trained by our proposed pre-training framework MMPT to downstream tasks. We use the lattice encoder  Latt$_{\rm{ENC}}$, and structure encoder to obtain the ${\bf{h}}_L$ and ${\bf{h}}_S$, respectively. Then, we utilize max-pooling and mean-pooling to the feature vector ${\bf{h}}_S$. Then, we concatenate ${\bf{h}}_L$ and features after the pooling operations  and  use four layers of the fully connected linear to obtain the final crystal property. When we train the eight downstream datasets, we set two types of learning rates. For the datasets aggregated from  MatBench Suite \cite{dunn2020benchmarking}, including JDFT2D, Dielectric and KVRH datasets, we use AdamW with a learning rate of 0.001 for training the parameters in the lattice encoder and structure encoder. We use a higher learning rate of 0.005 with the subsequence fully connected linear, which doesn't have the pre-training parameters in the pre-training model. For other datasets, we utilize AdamW by setting the learning rate as 0.01 with the OneCycleLR schedule strategy.

\subsection{Loss Function}
\label{lossf}
The total loss includes three parts, including reconstruction loss ${\cal{L}_{REC}}$, Barlow Twins loss ${\cal{L}_{BT}}$, and periodic attribute learning loss ${\cal{L}_{CAA}}$, which can be written as follows.
\begin{equation}
    {\cal{L}} = {\cal{L}_{REC}}+0.5\times{\cal{L}_{BT}}+{\cal{L}_{CAA}} = \alpha_1 \times {\cal{L}_{A}}+{\cal{L}_{X}}+{\cal{L}_{L}}+0.5 \times {\cal{L}_{BT}}+{\cal{L}}_{Die}+{\cal{L}}_{Unit}+{\cal{L}}_{Dis}.
\end{equation}
To facilitate atom type reconstruction, we compute the following loss:
\begin{equation}
    \begin{array}{l}
{\cal{L}_{A}} = \frac{1}{2}({\frac{1}{{|{\cal M}|}\times |\cal{B}|}\sum\limits_{b \in {\cal B}}\sum\limits_{i \in {\cal M}} {( - \sum\limits_{q = 1}^Q {{\hat {\bf{v}}_{b,i,q}}*} \log {{\bf{v}}_{b,i,q}}} ) + \frac{1}{{|\overline {\cal M|} \times |\cal{B}|}}\sum\limits_{b \in {\cal B}}\sum\limits_{i \in \overline {{\cal M}}} {( - \sum\limits_{q = 1}^Q {{\hat {\bf{v}}_{b,i,q}}*} \log {{\bar {\bf{v}}}_{b,i,q}}} )}),\\

\end{array}
\end{equation}

where  $\hat{{\bf{v}}}_{b,i,q}$  denotes the $q$-th element of the one-hot encoded atom type label for the $i$-th atom in the $b$-th crystal sample in a mini-batch.  $\alpha_1$ is set as 5 to emphasize the importance of atom type.
  Similarly, $\hat{{\bf{v}}}_{b,i,q}$ is the predicted 
  probability of atom type that $i$-th atom belongs to class $q$ as inferred from ${\bf{p}}_A$, whereas $\bar{{\bf{v}}}_{b,i,q}$ designates the predicted probability of atom type that $i$-th atom belongs to class $q$ as inferred from $\bar{{\bf{p}}}_A$.
$Q$ is the  total number of atom-type classes, which is set according to the chemistry element numbers. In our work, $Q$ is set as 119. $\cal{B}$ is the total indexes of the mini-batch.
Note that our framework only predicts the atom type with the mask token indices, and we averaged the cross entropy loss over masked atoms. Thus, our atom type loss function includes two parts that correspond to two mutex atom indices. 

\begin{equation}
\begin{split}
 {\cal{L}_{X}} = \frac{1}{2}(\frac{1}{{|{\cal M}| \times {|\cal B}|}} \sum\limits_{b \in {\cal B}}{ ||{{{d({{\bf{X}}^{b,P}},Mean({\bf{X}}^{b})) - d({{{\bf{X}}^{b,G}}},Mean({\bf{X}}^{b}))}}} ||_2}  \\+ \frac{1}{{|\overline{\cal M}| \times {|\cal B|}}} \sum\limits_{b \in {\cal B}}{ {{||{d({{\bf{\overline X}}^{b,P}},Mean({\bf{X}}^{b})) - d({\overline{{\bf{X}}}^{b,G}}, Mean({\bf{X}}^{b}))||}_2}} } ),
\end{split}
\end{equation}
where $b$ indexes batch samples. ${\bf{X}}^{b}$ is the coordinates for the $b$-th crystal sample in the mini-batch. ${{\bf{X}}^{b,P}}$ denotes the predicted coordinates for atom indices within subset $\cal{M}$, derived from ${\bf{p}}_C$, and this specifically pertains to the $b$-th crystal sample in the mini-batch.  ${{\bf{X}}^{b,G}}$ is the ground truth coordinates for atom indexes in subset $\cal{M}$. 
${{\bf{\overline{X}}}^{b,P}}$ is the predicted coordinates for atom indexes in subset $\overline{\cal{M}}$ according to ${\bar{\bf{p}}}_C$, and 
${{\overline{\bf{X}}}^{b,G}}$ is the coordinates  for atom indexes in subset $\overline{\cal{M}}$ of the $b$-th ground truth sample. $Mean({\bf{X}}^b)$ is the average of the whole atom coordinates for $b$-th crystal sample in the mini-batch. $d(\cdot)$ represents the distance utilized in \cite{xiecrystal}.

\begin{equation}
    {\cal{L}_{L}}= \frac{1}{6 \times |\cal{B}|}\sum\limits_{b \in {\cal B}}{||{\bf{L}}^b - {\bf{\hat L}}^b||_2^2},
\end{equation}
where ${\hat{\bf{L}}^b}$ is the predicted lattice for the $b$-th crystal sample in the mini-batch which is represented after Niggli Algorithm. The predicted lattice, which incorporates both directions and angles across three directions, consequently possesses six unique values.

Then, we introduce constraints for ${{\bf{p}}}_A^i$ and ${\overline{\bf{p}}}_A^i$  by calculating the Barlow Twins loss. We measure the cross-correlation matrix between two vectors ${{\bf{p}}}_A^i$ and ${\overline{\bf{p}}}_A^i$  and make it as close to the identity matrix as possible.

\begin{equation}
{{\cal{C}}_{ij}} \buildrel \Delta \over = \frac{{\sum\nolimits_b {{\bf{p}}_A^{b,i},{\bf{\bar p}}_A^{b,j}} }}{{\sqrt {\sum\nolimits_b {{{({\bf{p}}_A^{b,i})}^2}} } \sqrt {\sum\nolimits_b {{{({\bf{\bar p}}_A^{b,j})}^2}} } }},
{\cal{L}_{BT}} \buildrel \Delta \over = \sum\nolimits_i {{{(1 - {{\cal{C}}_{ii}})}^2} + \lambda } \sum\nolimits_i {\sum\nolimits_{j \ne i} {{{\cal{C}}_{ij}}} } ,
\end{equation}
where $b$ indexes batch samples and $i$,$j$ index the vector dimension of the networks’ outputs. $\cal{C}$ is a square matrix with the dimensionality of the network’s output. The Barlow Twins loss makes these two vectors to be similar while minimizing the redundancy between the components of these vectors.


We optimize the attribute prediction by calculating $\cal{L}_{CAA}$ loss, which includes ${\cal{L}}_{Die}$, ${\cal{L}}_{Unit}$  and ${\cal{L}}_{Dis}$ losses. 
\begin{equation}
\begin{split}
    {\cal{L}}_{Die} = \frac{1}{2}\frac{1}{{|{\cal{B}}| \times N \times n}}(\sum\limits_{b \in {\cal B}}\sum\limits_{j \in N \times n} {( - \sum\limits_{i = 1}^{27} {\hat {\bf{q}}_{i,j,b}^{Die}}  \times \log  {\bf{q}}_{i,j,b}^{Die}} )\\+\sum\limits_{b \in {\cal B}} \sum\limits_{j \in N \times n} {( - \sum\limits_{i = 1}^{27} {\hat {\bf{q}}_{i,j,b}^{Die}}  \times \log  \bar {\bf{q}}_{i,j,b}^{Die}} )),
    \end{split}
\end{equation}

\begin{equation}
\begin{split}
    {\cal{L}}_{Unit} = \frac{1}{2}\frac{1}{{|{\cal{B}}| \times N \times n}}(\sum\limits_{b \in {\cal B}}\sum\limits_{j \in N \times n} {( - \sum\limits_{i = 1}^2 {\hat {\bf{q}}_{i,j,b}^{Unit} \times } \log  {\bf{q}}_{i,j,b}^{Unit}} )\\+\sum\limits_{b \in {\cal B}}\sum\limits_{j \in N \times n} {( - \sum\limits_{i = 1}^2 {\hat {\bf{q}}_{i,j,b}^{Unit} \times } \log  \bar{{\bf{q}}}_{i,j,b}^{Unit}} )),
\end{split}
\end{equation}

\begin{equation}
\begin{split}
{\cal{L}}_{Dis} = \frac{1}{2}\frac{1}{{|{\cal{B}}| \times N \times n}}(\sum\limits_{b \in {\cal B}}\sum\limits_{j \in N \times n} {|\hat {{q}}_{j,b}^{Dis} -  {{q}}_{j,b}^{Dis}} |\\+\sum\limits_{b \in {\cal B}}\sum\limits_{j \in N \times n} {|\hat {{q}}_{j,b}^{Dis} -  \bar{{{q}}}_{j,b}^{Dis}} |) ,
\end{split}
\end{equation}

where $\hat {{\bf{q}}}_{i,j,b}^{Die}$ denotes the $i$-th element of the one-hot encoded ground truth, representing the discrete direction of the $j$-th edge in the $b$-th crystal sample within a mini-batch. ${\bf{q}}_{i,j,b}^{Die}$ represents the predicted probability of belonging to the discrete direction in the $i$-th class, as inferred from ${\bf{p}}_A$, whereas $\bar{{\bf{q}}}_{i,j,b}^{Die}$ designates the predicted discrete direction as inferred from $\bar{{\bf{p}}}_A$. $\hat {\bf{q}}_{i,j,b}^{Unit}$ is the $i$-th element of one-hot encoded ground truth regarding whether the atoms in the $j$-th edge belong to the same unit cell for the $b$-th crystal sample in mini-batch. $ {\bf{q}}_{i,j,b}^{Unit}$ is the predicted probability derived from $\bar{{\bf{p}}}_A$, and $ {{\bf{q}}}_{i,j,b}^{Unit}$ is the predicted probability based on $\bar{{\bf{p}}}_A$. Lastly, $\hat{{{q}}}_{j,b}^{Dis}$ denotes the ground truth of the distance between connected atom nodes in the $b$-th crystal sample in the mini-batch. ${{q}}_{j,b}^{Dis}$ is the predicted distance as determined by ${\bf{p}}_A$, while $\bar{{{q}}}_{j,b}^{Dis}$ represents the predicted distance as calculated from $\bar{\bf{{p}}}_A$.


\subsection{Baselines}
\label{baseline}
To evaluate the effectiveness of our framework, we compare our proposed framework with both supervised and self-supervised based baselines.
\begin{itemize}
    \item \textbf{CGCNN} \cite{xie2018crystal} is a crystal graph convolutional neural network that learns material properties from the connection of atoms in the crystal, providing a universal and interpretable representation of crystalline materials.
    \item  \textbf{SchNet} \cite{schutt2017schnet} is a novel deep-learning architecture modeling quantum interactions in molecules following fundamental quantum-chemical principles.
    \item \textbf{MEGNet} \cite{chen2019graph}  is a materials graph network with global state attributes for quantitative structure-state-property relationship prediction in materials.
    \item \textbf{ALIGNN} \cite{choudhary2021atomistic} is an Atomistic Line Graph Neural Network (ALIGNN), which performs message passing on both the interatomic bond graph and its line graph corresponding to bond angles.
    \item  \textbf{Matformer} \cite{yanperiodic} is a crystal material property prediction method based on periodic graph Transformers for encoding periodic structures of crystal materials.
    \item  \textbf{Crystal Twins} \cite{magar2022crystal} is a self-supervised learning framework for crystal property prediction based on graph neural networks.
    \item  \textbf{InfoGraph} \cite{Sun2020InfoGraph:} is an unsupervised and semi-supervised learning framework for graph representation based on graph neural networks. It maximizes the mutual information between graph-level representation and representation of substructures of different scales.
\end{itemize}

\subsection{Baseline Configuations}

In the following section, we show the configuration of the baseline models on the eight benchmark datasets we choose. We show the source of the results of the baselines that we directly report in the experiment results.

\textbf{CGCNN \cite{xie2018crystal}}: For the three MatBench \cite{dunn2020benchmarking} tasks, we report the results provided by the MatBench website. The model uses the original codes from Xie et al. \cite{xie2018crystal}, with 32 hidden dimensions, three CGCNN layers, and 128 as batch size. The training process uses early stopping if the validation loss does not improve for at least 500 epochs. The website doesn't specify the learning rate and training epochs. For other tasks, we report the results provided by Yan et al. \cite{yanperiodic}. The model has 128 hidden dimensions and three CGCNN message-passing layers. The model is trained for 1000 epochs,  the batch size is 256 and the initial learning rate is 1e-2.

\textbf{SchNet \cite{schutt2017schnet}}: For the three MatBench \cite{dunn2020benchmarking} tasks, we report the results provided by the MatBench website. The model is reimplemented in KGCNN \cite{reiser2021graph}, a package with several layer classes to build up graph convolution models. The model has four SchNet layers and the feature dimension is set as 64. The model is trained for 800 epochs with a batch size of 32. Adam optimizer and linear learning rating scheduler are used. For other tasks, we report the results provided by Yan et al. \cite{yanperiodic}. The model uses six layers of the SchNet message passing layer and a  feature dimension of 64. The model is trained with a learning rate of 5e-4 and a batch size of 64 for 500 epochs. The training uses an Adam optimizer with a 1e-5 weight decay and one cycle learning rate scheduler. In the task of MP\_Shear \cite{chen2019graph, jain2013materials} and MP\_Bulk \cite{chen2019graph, jain2013materials}, the 32-th smallest distance of a crystal is used to build the  graph. In the three rest tasks, the 12-th smallest distance of a crystal is used to build the graph.

\textbf{MEGNet} \cite{chen2019graph}:  For the three MatBench \cite{dunn2020benchmarking} tasks, we report the results provided by the MatBench website. The model is reimplemented in KGCNN \cite{reiser2021graph}. The model has three MEGNet layers and is trained for 1000 epochs with a batch size of 32. Adam optimizer and linear learning rating scheduler are used. For the other tasks, we report the results provided by Yan et al. \cite{yanperiodic}. The model uses three layers of the SchNet message passing layer and a  feature dimension of 64. The model is trained with a learning rate of 1e-3. The training phase uses an Adam optimizer with the OneCycleLR schedule strategy. Furthermore, the weight decay of the Adam optimizer is  set as  1e-5.  In the task of MP\_gap \cite{chen2019graph, jain2013materials}, a radius of 4.0 is used to build the graph, and the batch size is 128. For the rest tasks, the 12-th smallest distance of a crystal is used to build the graph, and the batch size is 64.

\textbf{ALIGNN \cite{choudhary2021atomistic}}: For the three MatBench \cite{dunn2020benchmarking} tasks, we report the results provided by the MatBench website. The model uses the original codes from Choudhary et al. \cite{choudhary2021atomistic}. For MP\_Shear \cite{chen2019graph, jain2013materials} and MP\_Bulk \cite{chen2019graph, jain2013materials}, we report the results provided by Yan et al. \cite{yanperiodic}. The model is trained with a 1e-3 learning rate and the batch size is set as 64. For other tasks, we report the results provided by Choudhary et al. \cite{choudhary2021atomistic}. 

\textbf{Matformer \cite{yanperiodic}}: For the three MatBench \cite{dunn2020benchmarking} tasks, we directly use the public code from Yan et al. \cite{yanperiodic}. Following the original paper \cite{yanperiodic}, we use a radius of 8.0 and the 12-th smallest distance between an atom and other atoms to build the multi-graph. We use a batch size of 64 to train the model. The training phase uses an Adam optimizer with the OneCycleLR schedule strategy.  For JDFT2D \cite{choudhary2017high, dunn2020benchmarking}, we train the model with a learning rate of 8e-4 for 300 epochs. For Dielectric \cite{jain2013materials, dunn2020benchmarking} and KVRH \cite{jain2013materials, de2015charting, dunn2020benchmarking}, we train the model with a learning rate of 1e-3 for 500 epochs. For the other five tasks, we report the results provided by the original paper \cite{yanperiodic}. The training phase uses an Adam optimizer with the OneCycleLR schedule strategy. For MP\_Shear \cite{chen2019graph, jain2013materials} and Jarvis\_gap \cite{choudhary2017high}, the model is trained with a learning rate 1e-3 for 300 epochs. For MP\_Bulk \cite{chen2019graph, jain2013materials} and Jarvis\_Ehull \cite{choudhary2017high}, the model is trained with a learning rate 1e-3 for 500 epochs. For Mp\_gap \cite{chen2019graph, jain2013materials}, the model is trained with a learning rate 5e-4 for 500 epochs.

\textbf{Crystal Twins \cite{magar2022crystal}}: For Crystal Twins, we directly use the public code for CT\_Barlow from Magar et al. \cite{magar2022crystal}. 
For the pre-training phase, we use a 128-dimension embedding encoder in the CGCNN. We use the Adam optimizer with a learning rate of 1e-5 and a weight decay of 1e-6. We use 80\% of the OQMD \cite{kirklin2015open, saal2013high, choudhary2020joint} dataset to pre-train the model, with a batch size of 64 for 20 epochs. For all the downstream tasks, we use the Adam optimizer with a learning rate of 1e-3 and a weight decay of 1e-6. Particularly, for MP\_gap \cite{chen2019graph, jain2013materials}, we train the model for 300 epochs with a batch size of 128. For all other downstream tasks, we train the model for 500 epochs with a batch size of 128.

\textbf{InfoGraph \cite{Sun2020InfoGraph:}}: Originally, InfoGraph use a 2D graph network \cite{xu2018powerful} as the encoder, and thus we change the encoder into a 3D graph network based on CGCNN \cite{xie2018crystal} encoder. We use a radius of 8.0 and the 12-th smallest distance between an atom and other atoms to build the crystal graph. For the pre-training phase, we use a 64-dimension embedding encoder in the CGCNN. We use the Adam optimizer with a learning rate of 1e-5 and a weight decay of 1e-6. We use 80\% of the OQMD \cite{kirklin2015open, saal2013high, choudhary2020joint} dataset to pre-train the model, with a batch size of 128 for 20 epochs. For all the downstream tasks, we use the Adam optimizer with a learning rate of 1e-3 and a weight decay of 1e-6. Particularly, for MP\_gap \cite{chen2019graph, jain2013materials}, we train the model for 300 epochs with a batch size of 256. For all other downstream tasks, we train the model for 500 epochs with a batch size of 256.
\subsection{Comparison with Self-Supervised Learning}

To demonstrate the need for SSL framework for crystal property prediction, we  use unsupervised InfoGraph \cite{Sun2020InfoGraph:} to pre-train a CGCNN \cite{xie2018crystal} encoder, and we compare the test MAE on the eight datasets.  Other SSL frameworks for non-crystal graph representation learning \cite{hu2019strategies, you2020graph, you2021graph, stark20223d, liu2022pretraining} aren't suitable for pre-training a crystal graph neural network for two reasons.
First, for learning with data augmentation methods \cite{hu2019strategies, you2020graph, you2021graph}, a periodic crystal graph cannot be directly augmented. Second, some molecular SSL frameworks \cite{stark20223d, liu2022pretraining}, require 2D molecular structure and chemical bond type information. However,  such details aren't typically accessible when examining the structure of a crystalline material. 

\begin{table}[!h]
\caption{Comparison between MMPT, supervised CGCNN, and InfoGraph in terms of test MAE on eight datasets.}
\renewcommand{\arraystretch}{1.1}
\centering
 \smallskip\resizebox{\columnwidth}{!}{
\begin{tabular}{c|c|c|c|c|c|c|c|c}
\toprule[1.6pt]
Method                                                                                & JDFT2D                     & Dielectric      & Mp\_Shear      & Mp\_bulk       & KVRH            & Jarvis\_gap & Jarvis\_ehull & Mp\_gap \\ \hline
\#   Crystals                                                                         & 636                        & 4764            & 5449           & 5450           & 10987           & 18171           & 55370         & 69239       \\ \hline

\begin{tabular}[c]{@{}c@{}}CGCNN\\  (2018, Phys. Rev. Lett.)\end{tabular}             & 49.2440                  & 0.5988          & 0.077          & 0.047          & 0.0712          & 0.41            & 0.170          & 0.292       \\ \hline

\begin{tabular}[c]{@{}c@{}}InfoGraph\\ (2020,ICLR)\end{tabular}             & 48.5135                      & 0.4684                          & 0.075                          & 0.046                         & 0.0674                      & 0.38                             & 0.128                              & 0.284   \\ \hline


MMPT                                                                                   & \textbf{38.2133} &  \textbf{0.3240} & \textbf{0.072} & \textbf{0.038} & \textbf{0.0567} & \textbf{0.26}   &   \textbf{0.061}            &    \textbf{ 0.204}        \\ \toprule[1.6pt]          \end{tabular}}
\label{result_infograph}
\vspace{-8pt}
\end{table}

As shown in Table \ref{result_infograph}, our purposed MMPT performs better than InfoGraph in all tasks. Considering the inability to acquire certain 2D information, we have opted to transition the encoder in InfoGraph to a CGCNN encoder.  Compared with CGCNN, InfoGraph improves the performance of CGCNN by a slight edge, which suggests that directly using a GNN pre-training framework to pre-train a crystal encoder is not enough.

\subsection{The Effectiveness of Multi-Graph Attention Mechanism}
We also conduct experiments to determine whether to apply our proposed multi-graph attention module to the structure encoder in the fine-tuning stage, as shown in Table \ref{fine}. MMPT-MGA is a variant without the Multi-graph attention mechanism.  The experimental results demonstrate that the multi-graph attention mechanism is beneficial during the fine-tuning process. 

\begin{table}[!h]

\centering
\caption{The ablation study in the fine-tuning stage.}
\smallskip\resizebox{0.6\columnwidth}{!}{
\begin{tabular}{c|c|c|c}
\toprule
Method   & Mp\_bulk       & Jarvis\_gap & Mp\_gap\\ \hline
MMPT-MGA & 0.040          & 0.28        &0.212   \\ \hline
MMPT     & \textbf{0.038} & \textbf{0.26}  & \textbf{0.204} \\ \toprule
\end{tabular}}

\label{fine}
\end{table} 
\end{document}